# Harmony Search: Current Studies and Uses on Healthcare Systems

Maryam T. Abdulkhaleq[1], Tarik A. Rashid[1], Abeer Alsadoon[2,3], Bryar A. Hassan[4], Mokhtar Mohammadi[5], Jaza M. Abdullah[6,7], Amit Chhabra[8], Sazan L. Ali[1], Rawshan N. Othman[1], Hadil A. Hasan[1], Sara Azad[1], Naz A. Mahmood[1], Sivan S. Abdalrahman[1], Hezha O. Rasul[9], Nebojsa Bacanin[10], S.Vimal[10],

[1]Department of Computer Science and Engineering, University of Kurdistan Hewler, KRI, F. R Iraq.

[2]School of Computing and Mathematics, Charles Sturt University, Sydney, Australia

[3]Asia Pacific International College (APIC), Information Technology Department, Sydney, Australia

[4]Department of Information Technology, Kurdistan Institution for Strategic Studies and Scientific Research, Sulaimani, 46001, Iraq

[5]Department of Information Technology, Lebanese French University, Erbil, KRI, F. R Iraq.

[6]Computer Science Department, College of Science, Komar University of Science and Technology, Sulaymaniyah, KRI, Iraq.

[7]Information Technology, College of Commerce, University of Sulaimani, Sulaymaniyah, KRI, Iraq.

[8]Department of Computer Engineering and Technology, Guru Nanak Dev University, Amritsar-INDIA

[9]Department of Pharmaceutical Chemistry, College of Medicals and Applied Sciences, Charmo University, Peshawa Street, Chamchamal, 46023, Sulaimani, Iraq

[10]Singidunum University, Danijelova 32, Belgrade, 11000, Serbia

[11]Dept of CSE Ramco Institute of Technology, Rajapalayam, Tamilnadu

Email (corresponding): bryar.hassan@kissr.edu.krd

**Abstract**

One of the popular metaheuristic search algorithms is Harmony Search (HS). It has been verified that HS can find solutions to optimization problems due to its balanced exploratory and convergence behavior and its simple and flexible structure. This capability makes the algorithm preferable to be applied in several real-world applications in various fields, including healthcare systems, different engineering fields, and computer science. The popularity of HS urges us to provide a comprehensive survey of the literature on HS and its variants on health systems, analyze its strengths and weaknesses, and suggest future research directions. In this review paper, the current studies and uses of harmony search are studied in four main domains. (i) The variants of HS, including its modifications and hybridization. (ii) Summary of the previous review works. (iii) Applications of HS in healthcare systems. (iv) And finally, an operational framework is proposed for the applications of HS in healthcare systems. The main contribution of this review is intended to provide a thorough examination of HS in healthcare systems while also serving as a valuable resource for prospective scholars who want to investigate or implement this method.

## 1. Background and Introduction

Before the year 2001, several search algorithms were already introduced and applied in different fields of applications. HS is one of the search algorithms that Zong Woo Geem has introduced. His vision was to develop a better-performing algorithm as an alternative to the other predefined algorithms. The intention of designing the algorithm is to have a generalized optimization technique for continuous, constrained, and discrete optimization in several types of optimization problems. Both the algorithm name and idea are inspired by the principle of the musician's improvisation; just like how musicians are always trying to choose the best and most potent harmony during their performances, this algorithm looks for the best solution (harmony) [1].

During the past years, the HS gained popularity, which can be observed through the diversified problems applied by the algorithm. HS's ability to solve various optimization problems attracted researchers and proved an effective optimization technique. Nowadays, the HS algorithm is still involved in engineering, computer science, industry, healthcare, construction, and robotics [2][3]. The factors that made the HS competitive with other algorithms are [3]:

- The ease of its mathematical model;
- No initial parameters are required to be set;
- The derivative information of the method is needless;
- The high flexibility of the algorithm;
- The fast and accurate nature of the method.

Many algorithms take the same path in finding inspiration to create their procedure ideas, such as Evolutionary Programming (EP), Particle Swarm Optimization (PSO), Evolution Strategies (ES), and Genetic Programming (GP), which are based on evolution. Some algorithms are based on some phenomena that can be seen in nature and the way living creatures think and act, such as Tabu Search (TS), Ant Colonies Optimization (ACO), and Neural Networks [2]. In contrast to the previously stated algorithms, the HS is based on an artificial phenomenon: the music improvisation process in which a given number of musicians seek to tune their instruments to achieve the best harmony (best state) [4][5].

In [6], Geem et al. developed the HS, a population-based meta-heuristic optimization method implemented in 2001. The HS mimics the construction of a unique harmony in musical composition to handle an optimization challenge. HS has been used in many complex problems, and it has produced outstanding results almost every time. Common examples of these problems are healthcare systems [7], fuzzy controllers and benchmark functions [8][9], beam and ball controllers [10], power flow analysis [11], congestion management [12], data mining [13], job shop scheduling [14], water distribution [6], university timetables [15], structural design, renewable energy [16], data clustering [17], and neural networks [18].

The HS's primary advantages are its clarity in implementation, its track record of success, and its capacity to deal with a wide range of complicated issues at once. The HS can make trade-offs between convergent and divergent areas, which is the primary reason for its vitality, success, and enviable reputation in the industry. Exploitation is mainly regulated by the pitch adjustment rate (PAR) and the bandwidth (BW) in the HS rule [6,19]. At the same time, exploration is primarily controlled by the harmony memory considering rate (HMCR) in the HS rule [20][21].

In addition, numerous survey papers and monographs [22] have discussed the advancements of primary HS and its variants. However, as far as we are aware, there is no comprehensive study of HS from its use in healthcare systems. A question raised here that how HS and its variants are utilized for the application domains of health systems. To address this question, this study aims to provide a comprehensive review of HS on healthcare systems to create a valuable source for researchers to inspire ideas for developing medical technologies. Therefore, the main contribution of this work is to examine the current research and applications of HS in healthcare systems.

This paper discusses HS and its extensions in healthcare applications four-fold. These are as follows:
1) HS variations, including alterations and hybridizations, are discussed in detail;
2) A synopsis of earlier review materials;
3) The applications of HS in the healthcare sector are reviewed;
4) The last section of the paper proposes an operational framework for using HS in healthcare systems.

As a critical contribution, this study aims to give a comprehensive assessment of the usage of HS in healthcare systems while also acting as a beneficial resource for potential researchers who want to explore or adopt this approach. In brief, we will make every effort to incorporate as many of these advancements as possible.

In this paper, we will start to present the current trends of HS in Section 2. In the next section, we will summarize the previsou review works on HS, followed by introducing the procedure of the original algorithm in Section 4. Next, we will discuss the most common modification that occurred to HS in Section 5; then, Section 6 will review the uses of HS in healthcare systems. In Section 7, we will propose an operational framework of HS variants to summarize their uses and procedures in healthcare problems. Finally, we will deliberate our conclusion and future trends in Section 8.

## 2. Current trends of HS

Since its first publication in a peer-reviewed journal, HS has established itself as a well-known population-based optimization method due to its broad adoption [23]. Since then, it has received widespread acceptance in various technical and medical sectors, as shown by many publications, as seen in Figure (1). As demonstrated by the number of papers on the technique's use since 2012 and the number of significant algorithm structure adjustments since 2006, there has been a sustained and rising interest in the HS and its variations [24]. Figure (2) depicts the number of issued articles, book chapters, conference papers, and other types of papers according to publisher types (ScienceDirect, IEEE, Springer, Wiley, and Taylor & Francis) since HS was proposed until December 2021. Figure (3) shows the number of published papers on each of the main variants of the HS, and Figure (4) presents the number of publications on applications of the HS to the healthcare sector.

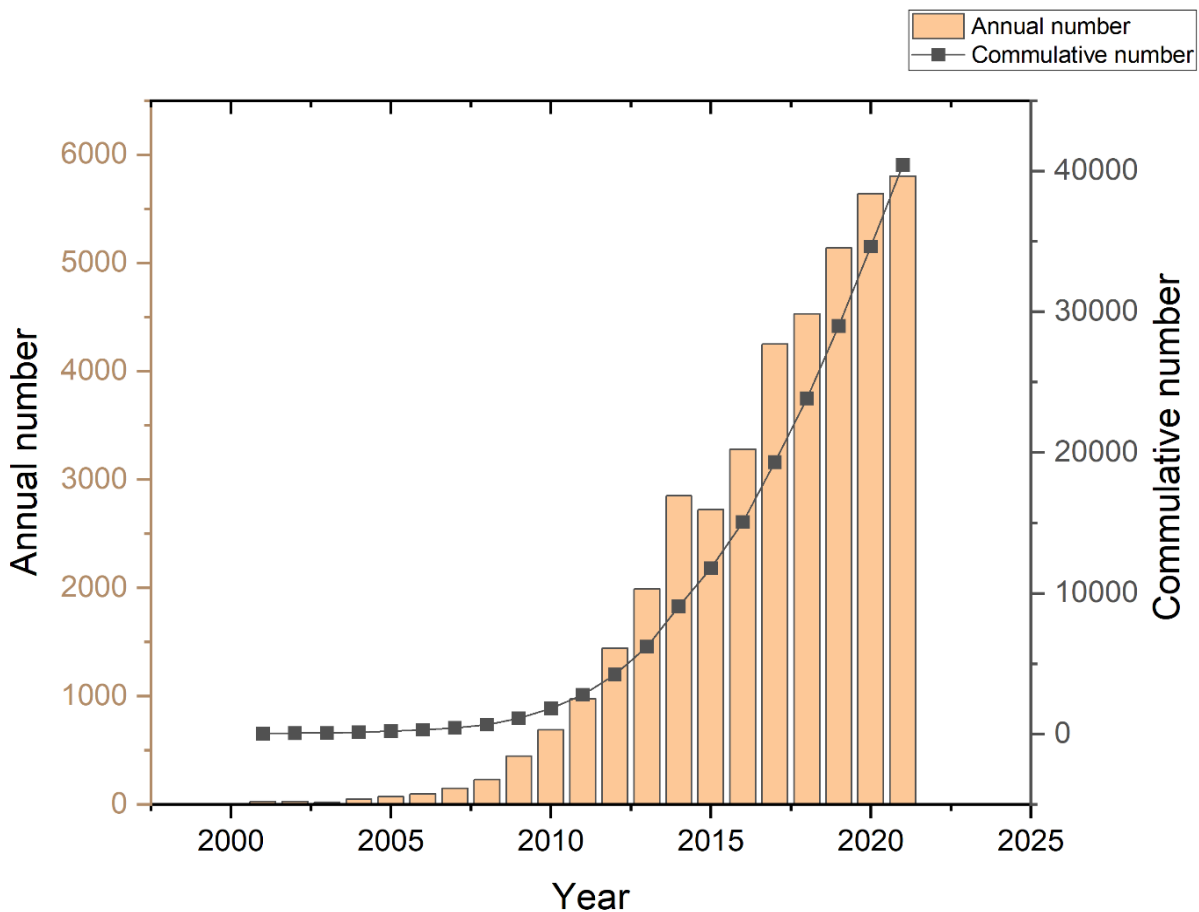

**Figure (1): Growing interest in the analysis and application of HS since 2001 (Source: Google Scholar; keyword search: "Harmony Search"; at 12: 40 Friday 31st December 2021)**

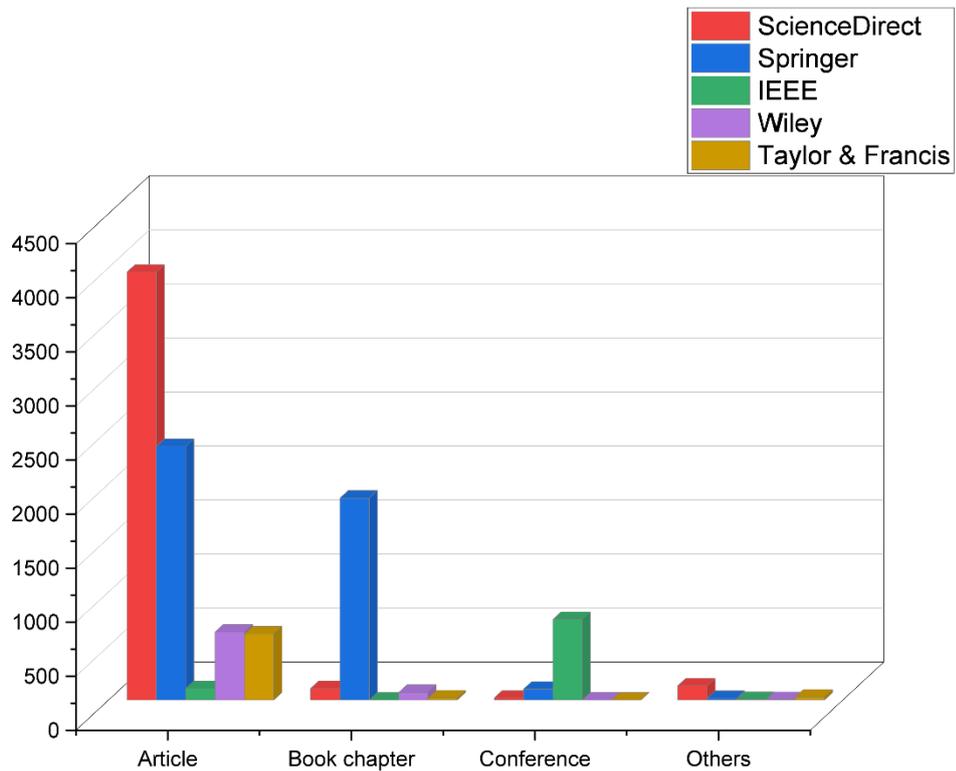

**Figure (2): Number of publications on the HS since it was proposed until December 2021, sorted by publisher and type (Source: ScienceDirect, Springer, IEEE, Wiley, and Taylor & Francis; keyword search: "Harmony Search"; at 12: 40 Friday 31st December 2021)**

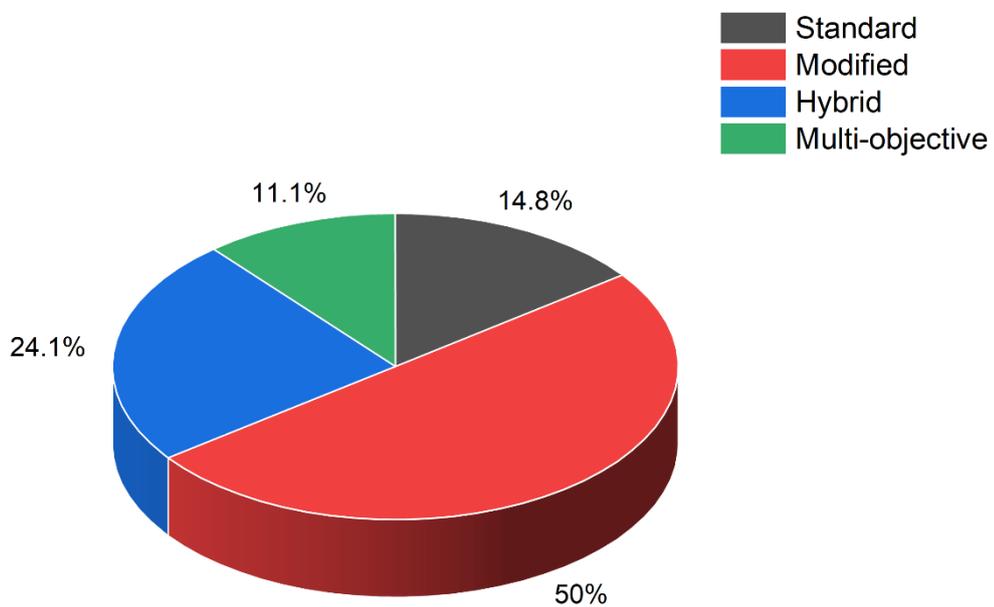

**Figure (3): Number of publications on variants of the HS since it was proposed until December 2021 (Source: Google Scholar, keyword search: "Harmony Search", at 12: 40 Friday 31st December 2021)**

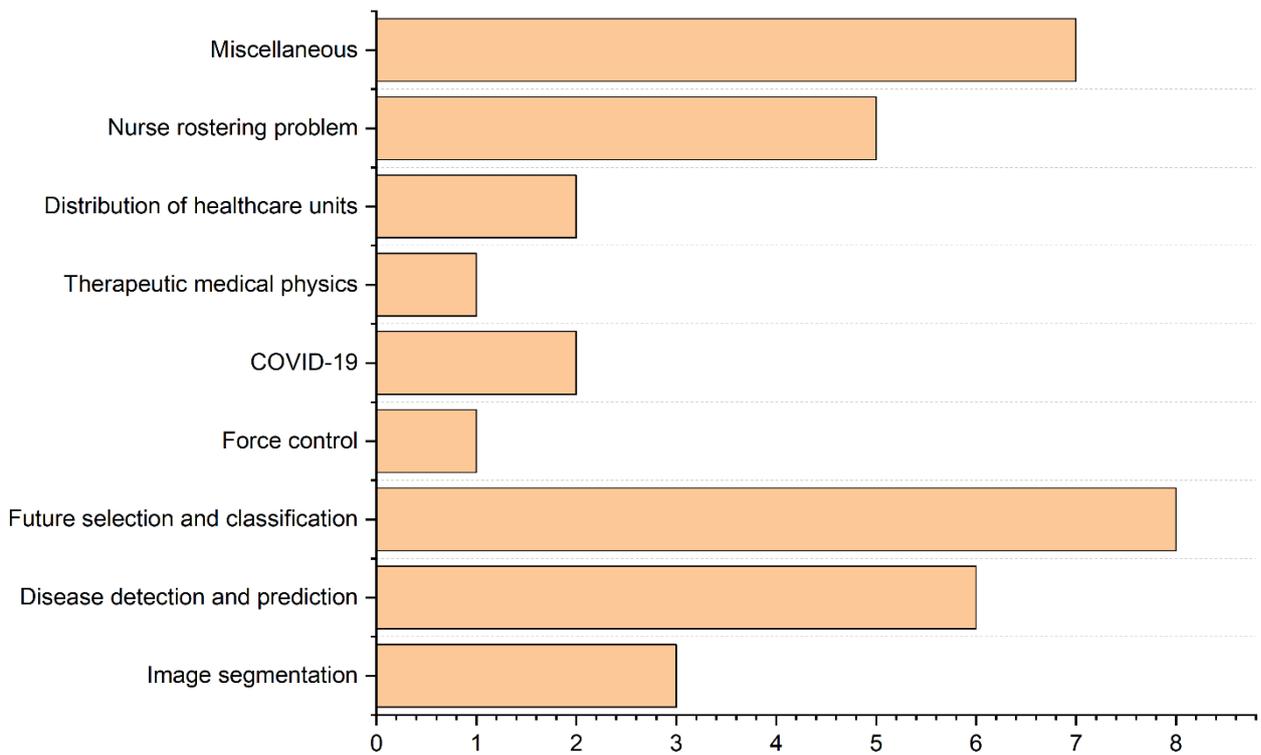

**Figure (4):** Number of publications on healthcare applications of the HS since it was proposed until December 2021 (at 12: 40 Friday 31st December 2021)

## 3. Summary of Previous Reviews

This Section briefs the recent study surveys in HS and outlines their achievements and identified research holes. Yang X-S. published a book chapter on HS in 2009 [25]. This chapter aims to review and discuss the novel HS in metaheuristic algorithms. He began by outlining the basic steps of HS and demonstrating how it operates. Next, he attempted to characterize metaheuristics and explain why HS is an effective metaheuristic algorithm. He then briefly discussed other common metaheuristics, such as particle swarm optimization, to compare and contrast them with HS. Finally, He explored strategies for improving and developing new HS varieties and making recommendations for further studies, including open questions.

In 2013, O. Abdel-Raouf and M. A.-B published an HS [1] survey to examine the distinctions between HS algorithms and their implementations. Additionally, some proposed potential enhancements are included, such as the deployment of HS on different applications. Manjarres D. et al. published another review article around the same time on the application of HS [26]. They comprehensively discussed and analyzed the main characteristics and implementation portfolio of HS. They claimed to review the most recent literature on the application of HS in several practical applications with the following three objectives: (i) to emphasize the excellent behavior of this modern meta-heuristic based on the increase in related contributions reported to date; (ii) to provide a bibliographic foundation for future research trends focusing on its applicability to other areas; and (iii) to do an in-depth study of potential possible research directions based on this meta-heuristic solver.

In 2016, Assad A. and Deep K. [13] published a survey report on the applications of HS in data mining. The analytical review was expected to provide an overview of the use of HS in data mining, particularly for those researchers interested in delving into the algorithm's capabilities in this area.

In late 2018, updated analysis and review were published by [27] on HS for energy engineering systems. The particular emphasis of this work had two primary objectives. To begin, the improved versions of HS adopted in recent studies are discussed. Second, contributions to energy system analysis using HS analyzed.

In 2019, four review works were published on HS with different objectives: (1) In early 2019, Ala'a A. et al. [28] published a comprehensive survey on the development of HS and its applications. The review paper discussed HS and its derivatives from a variety of perspectives. First, they explained the HS algorithm and demonstrated how its parameters influence its efficiency. Second, they defined HS classifications based on well-known HS variants and hybrid algorithms and the implementations of these classifications. Finally, a review of the HS algorithm's strengths and disadvantages and potential improvements was held. By highlighting similar studies from various disciplines, this article piqued curiosity about the implementation of HS for a broad range of audiences. (2) afterward, Yi J. et al. [29] published another survey work on HS's current state-of-the-art and its applications to intelligent manufacturing. This article offered a comprehensive overview of the HS fundamental principle and a survey of its most recent variants for feature optimization. Additionally, it included a study of cutting-edge HS applications in intelligent manufacturing, focusing on approximately 40 recently published papers. This paper also analyzed and summarized several possible future study avenues for HS and its applications to intelligent manufacturing. (3) Then, another review paper was published by Zhang T. and Geem Z. W. [22] to emphasize the historical evolution of algorithm structure rather than its implementations. This article discussed the initial HS in detail and various adapted and hybrid HS methods: adaptation of the basic HS operators, parameter adaptation, hybrid methods, solving multi-objective optimization issues, and restriction handling. (4) Finally, Yusup N. et al. published [29] a survey on HS focused on the feature selection method for classification. This article aimed to explore the efficiency of the HS for classification in various applications. The review concluded that HS function selection outperforms other well-known nature-inspired metaheuristics algorithms.

In 2020, Abualigah L. et al. [24] issued a comprehensive review of HS in various practical applications. In the same work, they reviewed papers in which HS was extended to multiple uses, most of which were clustering-related, such as data clustering, evidence clustering, fuzzy clustering, text clustering, wireless sensor networks, and image processing. Meanwhile, they discussed the strengths and shortcomings of HS and its variants to provide recommendations for potential studies.

Our review paper is distinct from previous work in several areas outlined below: (i) The prior review works are outdated, and they do not focus on specification applications of HS, such as healthcare systems. This review examines the most recent developments in various HS on healthcare systems. (ii) The implementation of HS is explained by giving the step-by-step procedure of HS and a practical example. (iii) The state-of-the-art reviews on HS and its variants are also discussed. (iv) This article delves deeply into healthcare environments where HS is widely employed and highlights notable work to inspire scholars in the field of meta-heuristic algorithms. (v) An operational framework of HS is proposed to review the leading implementation of HS and its extensions on healthcare systems and find the relationships among these variants. (vi) It illustrates the challenges and offers alternative solutions. We discovered about 300 latest study articles using Google Scholar by feeding various variations of the term 'harmony search, HS, harmony search algorithm' into the search box. After critically analyzing the collected documents, we identified around 60 as the most important in HS and its variants on healthcare systems, which are addressed in this survey.

## 4. The Standard HS

Harmony composition in terms of music is when the musicians attempt to generate new harmony by combining notes emitted in sequence. The composition process changes the harmonies seeking improvement until they get a satisfactory result [5][30]. They made formalization on the three measurable optimization options, which are [31]:

1) Select and play any popular from the musician's memory;
2) Playing something similar to a tune that has been played before;
3) Composing some new notes, which can be a random note.

Following the same concept in HS, we will have three components that the HS utilizes to have an equivalent process to the harmony composition that we mentioned before. These components are HM, pitch adjusting, and randomization [31]. Also, it is noteworthy to note that there are two parameter rates in HS [32]:

1) HM considering rate (HMCR): - this is used to define the set value (1.0-HMRC) in the HM (HM);

2) Pitch adjusting rate (PAR): this parameter is used only after choosing a value from the HM because it sets the pace of regulating the pitch selected from the HM (1-PAR).

These components and parameters are used in the HS variable decision where a value that follows one of the three rules as the following [21,30–33] :

1) HM: selecting any single weight from the HM;
2) Pitch adjusting: determining an adjacent value of one value from its memory;
3) Randomization: choosing a random value from the possible values range.

The HS has the following five steps [21,31–38]:

<u>Step 1:</u> initializing the parameters of the algorithm and the optimization problem, which is represented in Equation (1) as follows:

$$\text{Min}\{f(x)|\ x_i \in X_i\}, \ i = 1,2,\dots,N \tag{1}$$

Where:
$f(x)$: The objective function.
$x_i$: decision variables set.
$X_i$: The possible range of values for each decision variable ($x_i$) that is $X_i \in [L^{x_i}, U^{x_i}]$ where $L^{x_i}$ And $U^{x_i}$ are the minimum and maximum bound for $x_i$.
$N$: Number of decision variables.

In this step, the control parameters are quantified. These parameters are:

(1) HM Size (HMS): this parameter can be considered the equivalent to the size of the population in population-based algorithms;
(2) The HM Consideration Rate: This is used to check if the decision variable value is selected from the stored solution in the HM;
(3) The Pitch Adjustment Rate: this parameter determines whether or not the decision variables have been changed to a neighboring value;
(4) The Distance Bandwidth (BW) value specifies the adjustment distance for the pitch adjustment operator;
(5) The Number of Improvisations (NI).

<u>Step 2:</u> Initialize the HM; in this step, the HM is generated by filling the matrix with randomly generated solution vectors, as shown in Equation (3). The numbers are between the maximum and the minimum bound $[L^{x_i}, U^{x_i}]$ where $1 \leq i \leq N$. The values are generated by Equations (2) and (3).

$$x_i^j = L^{x_i} + rand(0,1) \times (U^{x_i} - L^{x_i}) \text{ where } j = 1,2,3, \ldots HMS, \tag{2}$$

$$HM = \begin{bmatrix} x_1^1 & x_2^1 & \ldots & x_N^1 \\ x_1^2 & x_2^2 & \ldots & x_H^2 \\ \ldots & \ldots & \ldots & \ldots \\ x_1^{HMS} & x_2^{HMS} & \ldots & x_N^{HMS} \end{bmatrix} \tag{3}$$

<u>Step 3:</u> new harmony improvising from the HM, In this step, a new harmony vector is generated $\vec{x} = (\acute{x}_1, \acute{x}_2, \acute{x}_3, \ldots, \acute{x}_N,)$ based on the three rules:

1) Harmony Memory (HM): The first variable decision $\acute{x}_1$, is selected from the HM range $[\acute{x}_1 - x_1^{HMS}]$ with probability (w.p): HMCR where HRMS $\in (0,1)$;
2) Pitch adjusting: using PAR parameter: Each decision variable is examined to check if it should be pitch-adjusted with probability PAR, where PAR $\in (0,1)$ is presented in Equation (4).

$$\text{Pitch adjusting decision for } \acute{x}_1 \leftarrow \begin{cases} \acute{x}_1 = \pm rand(0,1) \times BW & w.p \quad PAR, \\ \acute{x}_1 & w.p \quad 1 - PAR \end{cases} \tag{4}$$

3) Randomization: the first variable decision $\acute{x}_1$, s selected from a random value that already exists in the current HS with a probability of $(1 - HRMS)$ as it is shown in Equation (5)

$$\acute{x}_1 \leftarrow \begin{cases} \acute{x}_1 \in \{x_i^1, x_i^2, x_i^3 \ldots, x_i^{HMS}\} & w.p \, HMRC, \\ \acute{x}_1 \in X_i & w.p \, 1 - HRMS \end{cases} \tag{5}$$

<u>Step 4:</u> Update the HM, checking if the New Harmony vector $\vec{x} = (\acute{x}_1, \acute{x}_2, \acute{x}_3, \ldots, \acute{x}_N,)$ is better than the worst harmony in the HM; the HM will be updated by excluding the worst harmony and including the best one.

<u>Step 5:</u> Repeat the previous steps 3 and 4 to satisfy the termination criterion.

The pseudocode of the HS algorithm is presented in the algorithm (1).

1. Begin
2. Define objective function $\{f(x), \backslash x=(x), X2 \backslash .....Xa)\}^{\wedge}T$
3. Define (HMCR), (PAR) and the other parameters
4. Generate (HM) with random harmonies
5. While (t<max number of iterations)
6.    While (i<=number of variables)
7.       If (rand<HMCR),
8.          Choose a value from HM for the variable i
9.          If (rand<PAR).
10.            Adjust the value by adding a certain amount
11.          End if
12.       Else
13.          Choose a random value
14.       End if
15.    End while
16.    Accept the new harmony (solution) if better
17. End while
18. Find the current best solution
19. End

**Algorithm (1): Pseudocode of HS**

## 5. Variants of HS

Due to the popularity of the HS, various publications have been proposed, and some of them are applied in different applications to achieve better performance than the original HS [1]. Scholars were inspired to enhance the standard HS's efficiency due to its alleged deficiencies in respect to deficiency and convergence speed. Numerous scholars have proposed modifications to the original HS algorithm, although others have adapted it to several issues. This Section discusses the developments of HS variants and the current published work on them. These variants are classified into two main categories: modifications and hybridizations of HS. Figure (5) depicts a general framework of the main variants of HS.

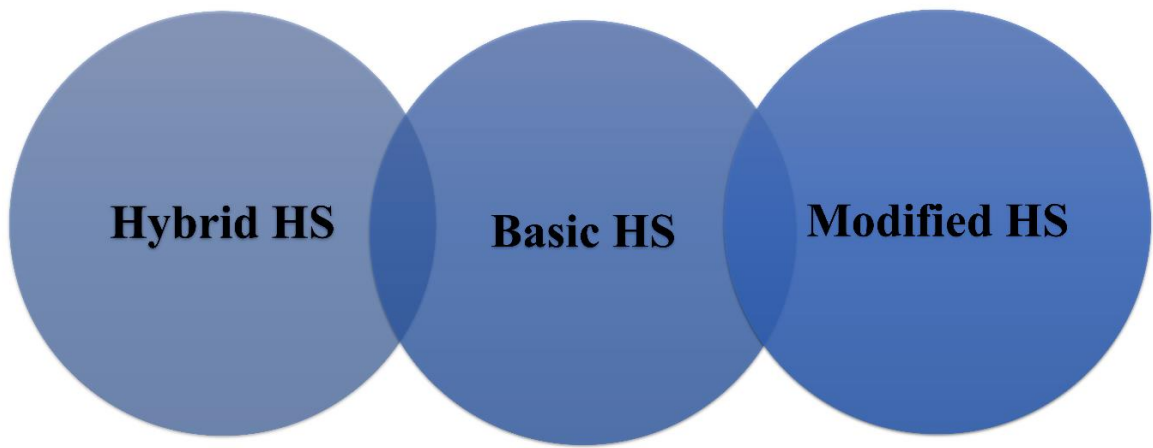

**Figure (5): Variants of HS**

### 5.1. Modifications of HS

Scholars have updated it numerous times to enhance HS's efficacy and efficiency and make it more suitable for a particular application. This Section discusses five adjusted HS algorithms. The modified algorithms are HIS, GHS, NGHS, SAHS, and CHS. These algorithms are summarized in Table (1).

**Table (1): The modified HS algorithms**

| Algorithm | Reference, author(s) | Aim | Modification method | Performance index | Concluding remarks |
|---|---|---|---|---|---|
| iHS | [39], M. Mahdavi et al | Solving practical problems | Tuning HS constant parameters | iHS > HS | iHS works better than HS and other heuristic methods and is an excellent search for many engineering optimization problems. |
| GHS | [40], M. G. H. Omran and M. Mahdavi | Improving the performance of HS | Borrowing swarm intelligence principles improve the efficiency of HS | HS performance was improved | GHS seems to be a viable solution for solving problems, including Integer Programming. |
| NGHS | [41], D. Zou et al. | Solving reliability problems | Performing two critical operations: location updating and low-probability genetic mutation | NGHS | NGHS can be a viable option for resolving reliability issues. |
| SAHS | [38], C.-M. Wang and Y.-F. Huang | Tuning appropriate parameter values | Adjusting automatic parameter values | SAHS | SAHS has supremacy over the original HS and its newly evolved derivatives. |
| IHSCH | [42], O. Abdel-Raouf et al. | Solving assignments | Generating a candidate solution by disorderly action is comparable to acoustic monophony. | IHSCH > HS and IHS | IHSCH is a more reliable and more effective process than conventional approaches (HS and HIS). |

**5.1.1. IHS.** The objective of IHS is to enhance the performance by evaluating both pitch adjusting rate and distance bandwidth values because small values of PAR and large values of BW can reduce the performance of this method [43][39]. In the Original HS (OHS), these parameters were fixed [44], which made Mahdavi et al. add their modifications in 2007 [5] by applying a dynamic technique to archive their values dynamically [28]. The dynamic PAR is shown in Equation (6) as follows:

$$\text{PAR}(N) = \text{PAR}_{Min} + \frac{(\text{PAR}_{Max} - \text{PAR}_{Min})}{NI} * n. \tag{6}$$

Where [40]:

PAR(N) : the PAR value for n generation.

$\text{PAR}_{Max}$: maximum adjustment value.

$\text{PAR}_{Min}$ maximum adjustment value.

Furthermore, the update of the BW will be based on the Bwmax and Bwmin value of BW as follows in Equation (7) [40]:

$$BW(N) = bw_{max} \, Exp\left(\left(\frac{Ln\left(\frac{BW_{Min}}{BW_{Max}}\right)}{NI}\right) * n\right) \tag{7}$$

Where:

$BW(N)$ = the bandwidth of the generation n.

Bwmax = the maximum BW value.

Bwmin = The minimum BW value.

The IHS was applied to solve many complex problems such as (minimization of the weight of spring, welded beam design disjoints feasible region, etc.) [39]. It is also used to solve the distribution network problems [45]. Consequently, it provides better performance than the OHS regarding noise and high dimensions [1]. Similarly, IHS generated a new solution vector that increases the accuracy and enhancement rate of the HS [5][46]. However, the main issue of IHS is the difficulty in choosing the minimum and maximum values of the bandwidth within a wide range (0-infinity) [25]. Similarly, the values of bwmax and bwmin can be introduced to other problems by determining and guessing them by the user because they are not independent [47].

**5.1.2. GHS.** After the presence of the IHS, M. Omran and M. Mahdavi proposed another modified algorithm called the Global-best HS (GHS) while they were trying to compare their study of applying the IHS algorithm to designing water-network problems with the other stochastic algorithms [28,47]. The authors noticed the limitations of IHS, which has the struggle of finding the upper and lower bandwidth bounds (BW). Therefore, they worked to overcome this problem by integrating the concept of PSO. The PSO in GHA deals with particles defined as a swarm of individuals where each of these particles represents a candidate solution. The only difference between the HS and GHS is this approach, which changes the pitch-adjusting step whereby the BW is replaced with the best HM value. This step adds a more social dimension to the algorithm. Equation (8) demonstrated how GHS utilizes the excellent result in HM to substitute BW values [40].

If $rand(0,1) \leq PAR(t)$, then Produce a random number $k \in U(1,..D)$ (8)

The GHS gives more successful results than HS and IHS concerning high dimensions and noise problems. It is the best choice to apply in applications, such as water network design rather than HS in minor and medium-scale issues; however, the GHS performance is worse in large-scale problems than HS [40].

**5.1.3. NGHS.** In 2010, Zou et al. Intended to update GHS to make it able to solve complex reliability problems [41]. They proposed a modified algorithm called Novel NGHS, which has two additional operations: position updating and genetic. The advantage of NGHS is its ability to turn the worst harmony in the HM into a global best harmony in each iteration rapidly. However, convergence problems may appear; therefore, the genetic mutation operator should solve this problem [48][49]. The pseudocode of NGHS is available in [41].

**5.1.4. SAHS.** The authors of [38] proposed a modified algorithm in 2010 that is also concerned with the pitch adjustment since the BW and PAR parameters impact the value of the final solution. Their proposed method was the parameters self-adapting based on the best experience. The concept of the randomized number is replaced with the low-discrepancy sequences in the HS initialization represented by Equations (9) and (10) so that the latest harmony is revised by the HM's maximum and minimum values.

$trial^i + [\max{(HM^i - trila^i)}] \times rand[0,1]$ (9)

$trial^i - \min{(HM)^i}] \times rand[0,1]$ (10)

Where:

$trial^i$ = the pitch adjusted that is chosen from the HM.

$\text{Max}(HM^i)$ & $\min{(HM^i)}$ = the maximum and minimum values of the HM [38].

**5.1.5. IHSCH.** The improved version of a harmony metaheuristic algorithm with different chaotic maps (IHSCH) is another version of HS used to solve linear assignment problems, considered one of the main optimization problems. This problem can be found in many scheduling applications such as planes and crews. These kinds of applications have many tasks and agents, and each agent should be assigned to one task so that the total assignment cost is reduced. This problem can be presented in Equation (11).

$$Min\ f(x) = \sum_{i=1}^{n}\sum_{j=1}^{n} c_{ij}\ x_{ij} \qquad (11)$$

Subject to:

$$\sum_{j=1}^{n} x_{ij} = 1,\ i = 1,2,3,\dots,n \qquad (12)$$

$$\sum_{i=1}^{n} x_{ij} = 1,\ j = 1,2,3,\dots,n \qquad (13)$$

An improvement has been presented in the HS algorithm to overcome this problem, embedding the nature of chaos with HS to find a solution. Testing the algorithm with three problems takes less time than traditional algorithms, and optimal solutions are obtained. The modified algorithms' foremost step is to generate long and consistent sequences to simulate complex problems quickly. After testing the modified algorithm with different linear assignment problems, the results show that total time is reduced beside the total cost [42].

### 5.2. Hybridizations of HS

Hybridization is a widespread and successful technique to improve an algorithm. Hybrid algorithms combine or merge two existing algorithms, resulting in a new and robust algorithm with enhanced capabilities. The hybrid HS algorithms have proved their efficiency in performance optimization to a higher level [28]. This enhancement can be represented in global search ability and/or exploration, the quality of the solutions, computational cost, and convergence speed. Several types of research used the concept of merging other algorithms with HS. However, we can categorize this hybridization into two main categories [50]:

1) The first category is the incorporation of HS with algorithms: In this case, the researchers will attempt to improve the HS by exploiting the advantages of another algorithm or using it to overcome some drawbacks in HS [50]. For instance, Niu et al. borrowed the arithmetic crossover operation from GA to increase the range of HS solutions, which will solve the premature convergence problem [51].
2) The second category is incorporating other algorithms into HS: in such cases, the HS is used as a component to improve the different algorithms, just like how Kaveh et al. introduced a new hybrid algorithm (HPSACO) based on PSOPC, ACO, and HS. They employed HS to control variable constraints in truss structure design [28][52].

## 6. Applications of HS in healthcare systems

A comprehensive overview of the HS and its variants was explained in the previous sections. The HS algorithm is an optimization algorithm for comparing solutions to achieve the optimum result for issues in different applications, including the biological and medical fields. There are many problems in this area that the HS algorithm can be used to determine its optimal result, especially since the most essential characteristics of this algorithm are its simplicity and ease of code and its fast results. In this section, we will examine what medical problems HS can be utilized by giving ideas that have been proposed in different academic publications. HS's use in healthcare systems includes image segmentation, disease detection and prediction, future selection and classification, force control, COVID-19, therapeutic medical physics, distribution of healthcare units, rostering and scheduling problems, and miscellaneous. The applicability of HS and its variants in different research domains of healthcare systems are shown in Figure (6).

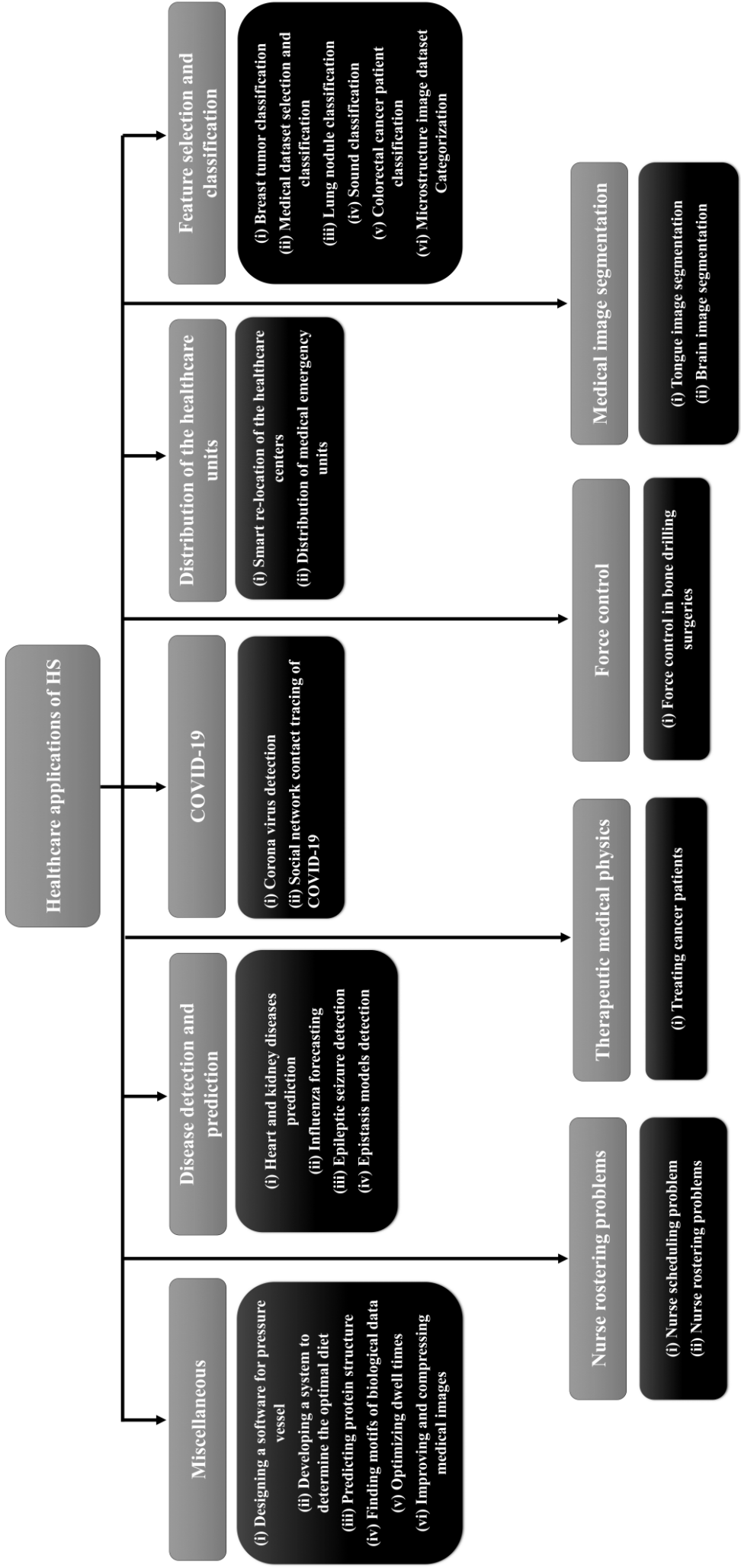

**Figure (6): Applications of HS and its variants in the healthcare domain**

## 6.1. Medical image segmentation

One of the applications of HS is to find the optimum value of clustering numbers in medical image segmentation. Table (2) summarizes the applications of HS and its variants in medical image segmentation.

Table (2): Applications of HS and its variants in medical image segmentation

| Reference, author(s) | Proposed algorithm | Aim | Performance index | Concluding remarks |
|---|---|---|---|---|
| [53], Srikanth R, and Bikshalu K | The proposed algorithm | Image segmentation of the tongue | The proposed algorithm was more efficient than the other compared techniques | The proposed method is superior to histogram-based methods. |
| [54], Fachrurrozi M et al. | Hybrid multilevel thresholding and HS (HSA) | Image segmentation of the tongue | HSA (medium level parameter) | The minimal distance classifier, the k-NN classifier, and the SVM classifier are employed for classification, and HHS achieves an average classification accuracy of 98.19 percent, 98.34 percent, and 97.18 percent, respectively, with the fewest features. |
| [55], M. S. H. Lipu et al. | Improved HS (HIS) | Magnetic resonance imaging (MRI) brain image segmentation | IHS | IHS was superior to the fuzzy clustering algorithm for MRI image segmentation. |

## 6.2. Disease detection and prediction

The use of HS for detecting and predicting diseases has gained more importance in the medical decision support systems. Referring to Table (3), HS has been used to predict, detect, and diagnose several diseases.

**Table (3): Applications of HS and its variants in disease detection and prediction**

| Reference, author(s) | Proposed algorithm | Aim | Performance index | Concluding remarks |
|---|---|---|---|---|
| [56], Tuo S et al. | MP-HS-DHSI | Detecting high-order single nucleotide polymorphism (SNP) interactions | MP-HS-DHSI outperformed four state-of-the-art algorithms | The suggested strategy may significantly be improved the search speed and discriminating ability of a variety of epistasis models |
| [57], Koti P et al. | Hybrid HS (HM-L) method combined with the Levi distribution | Predicting kidney diseases | HM-L algorithm outperformed other state-of-the-art methods | Sensitivity of 96, specificity of 93.33, the accuracy of 95, F-score of 96, and kappa value of 0.89 were obtained with the provided HM-L model |
| [58], Tuo S et al. | Niche HS | Detecting complex disease-associated high-order SNP combinations | Niche HS outperformed several traditional algorithms in detection power and CPU runtime for all of these datasets. | Niche HS showed promise in discovering high-order disease-causing models when applied to age-related macular degeneration (AMD). |
| [59], Hickmann KS et al. | The forecasting model | Forecasting the 2013-2014 influenza season | N/A | For the 2013-2014 ILI observations, the forecasting approach yielded 50 percent and 95 percent credible intervals that encompassed the actual observations for most weeks in the prediction. |
| [60], Gandhi TK et al. | DHS-MD | Detecting epileptic seizure activity with | DHS-MD compared to others | DHS-MD detection was 100% accurate with the same degree |

| | | very fast and highest accuracy | | of sensitivity and specificity. |
|---|---|---|---|---|
| [61], P. Koti et al. | Hybrid HS with Levi Distribution (HM-L) | Predicting heart diseases | Hybrid HS | The accuracy of disease diagnosis was improved with less time. |

## 6.3. Feature selection and classification

HS has been used for several medical classification tasks. Table (4) reviews the use of HS in different feature selection and classification applications.

**Table (4): Applications of HS and its variants in feature selection and classification**

| Reference, author(s) | Algorithm | Aim | Performance index | Concluding remarks |
|---|---|---|---|---|
| [62], Sarkar SS et al. | Proposed HS | Categorizing microstructure image datasets | Proposed HS | HS method achieved a higher agreement between feature selection and classification accuracy than the other methods. |
| [63], Dash R. | Adaptive HS | Gene selection and classification of high dimensional medical data | Adaptive HS | Adaptive HS was competent compared to other considered approaches. |
| [64], Bae JH et al. | Z-FS-KM-MHS | Classifying colorectal cancer patients from healthy people | Z-FS-KM-MHS was more accurate than the other compared methods | The suggested model achieved an accuracy of up to 94.36 percent in classifying objects. |
| [65], Mousavi SM et al. | OFHS | Classifying medical datasets | OFHS was more efficient than the other compared algorithms | OFHS was robust in data analysis and classification of clinical datasets. |
| [66], S. Kar et al. | Group improvised HS (GrIHS) | Lung nodule classification | GrIHS had a sensitivity of 97.59% and accuracy of 97.78% | Provision of high accuracy for ling nodule blind testing. |

| [67], Sudha MN, and Selvarajan S. | Hybrid of cuckoo search and (HHS) | breast tumor classification | HHS was more accurate than a genetic algorithm (GA) and particle swarm algorithm (PSO) | The minimal distance, the k-NN, and the SVM classifier were employed for classification. With the fewest features, HHS achieved an average classification accuracy of 98.19 percent, 98.34 percent, and 97.18 percent, respectively. |
|---|---|---|---|---|
| [68], E. Alexandre et al. | HS | Sound classification | HS | HS has outperformed its competitors in respect of error rate and execution time. |
| [69], Alexandre E et al. | Music-inspired HS | Classifying sounds in hearing aids | Music-inspired HS | Music-inspired HS was compared with sequential search algorithms or random search on 74 different features to test its performance. |

## 6.4. Force control in bone drilling surgeries

For several years ago, bone drilling in surgeries has been used and different surgeries. This requires the prediction of increasing and minimizing the thrust force. Vibration-Assisted Drilling (VAD) was presented [71] to achieve this aim. This research used IFSHS. In the same study, a drilling algorithm developed results in an optimal cutting state. The optimal solution is achieved by applying evaluation information feedback, optimization parameters, and dynamic searching parameters. Experiments showed that VAD reduces the force by 18.4% to 33.2% compared to CD. Compared to the original VAD, the drilling force decreased by 20.19%. The way the HS algorithm works has been presented previously in this work. IFSHS operates using self-adaptive parameters that consist of a dynamic parameter and feedback factors.

## 6.5. COVID-19

HS and its variants have been used for different purposes in COVID-19, such as image enhancement for Coronavirus detection and social network contact tracing. Table (5) recaps the use of HS in Coronavirus.

Table (5): Applications of HS and its variants in COVID-19

| Reference, author(s) | Proposed algorithm | Aim | Performance index | Concluding remarks |
|---|---|---|---|---|
| [70], Al-Shaikh A et al. | Hybrid harmony search (HHS) | Social network contact tracing of COVID-19 | HHS showed its superiority over these algorithms | HHS improved run time by 77.18 percent and had an excellent average error rate of 1.7 percent compared to other current algorithms for locating tightly coupled components (SCCs). |
| [71], V. Rajinikanth et al. | HS and Otsu based System | COVID-19 disease detection | The proposed method extracted the diseased portion more precisely | HS was beneficial in preventing the diagnostic burden associated with mass screening processing. |

## 6.6. Therapeutic medical physics

Medical physics is almost as essential as medicine because it involves radiation for medicinal and diagnostic purposes. This analysis compared HS and GA in optimization simulations for HDR brachytherapy for prostate cancer, where the optimum treatment strategy is determined by analyzing radiation dose-based optimization constraints. Since therapeutic radiation plays a critical role in cancer care, it is possible to use optimization techniques such as HS to provide a hefty dose of radiation to cancer cells in this situation. The HS has several benefits that make it ideal for these uses, one of which is that it is quicker than the GA, which is vital for patients to improve in medical physics. Additionally, the HS can be highly effective in HDR prostate brachytherapy and is well-suited for mixing its parameters with optimum parameters to achieve a considerably shorter treatment period. This capability is critical for time-consuming clinic procedures such as IMRT, tomotherapy, brachytherapy, and beam angle optimization. The simulation environment was constructed using various methods, including wxpython, Matplotlib, NumPy, and Python. - of these resources is free and open-source [72].

### 6.7. Distribution of the healthcare units

The two major problems are targeted, such as smart re-location and distribution of healthcare units. Table (6) presents the two main applications of HS and its variants in the distribution of healthcare units.

Table (6): Applications of HS and its variants in the distribution of healthcare units

| Reference, author(s) | Proposed algorithm | Aim | Performance index | Concluding remarks |
|---|---|---|---|---|
| [73], M. Alinaghian et al. | CPLEX | Locating the healthcare centers based on their facilities | CPLEX | CPLEX provided a reasonable solution for locating healthcare centers based on their facilities. |
| [74], Landa-Torres I et al. | Multi-objective grouping HS (MOGHS) | Distribution of medical emergency units | MOGHS | The encouraging findings open the way for the proposed technique to be used in real-world circumstances. |

### 6.8. Rostering and scheduling problems

The Nurse Rostering Problems (NRPs) is a management study for finding an ideal way for scheduling nurses for their shift and patient admission with a set of restraints. The uses of HS and its variants in rostering and scheduling problems are reviewed in Table (7).

Table (7): Applications of HS and its variants in rostering and scheduling problems

| Reference, author(s) | Proposed algorithm | Aim | Performance index | Concluding remarks |
|---|---|---|---|---|
| [75], Doush IA et al. | Novel HS | Patient admission scheduling (PAS) | The novel HS algorithm was capable of producing comparably competitive results when compared with nine state-of-the-art algorithms | The suggested HS was a highly efficient solution to the PAS problem that may address various scheduling issues involving enormous amounts of data. |

| [76], Yagmur EC, and Sarucan A. | Opposition-based parallel HSA | Nurse scheduling problem | Opposition-based parallel HSA > HS | The proposed approach had achieved better results than the standard HS in all eight different situations |
|---|---|---|---|---|
| [77], Hadwan M. et al. | Enhanced HS with great deluge algorithm (GD), called DHSA | Solving nurse rostering problems | DHSA > the hybridization of enhanced HS with hill climbing (HC) (CHSA) | DHSA performed much better than CHSA in solution quality, with a slightly higher execution time in all instances. |
| [78], Hadwan M et al. | CHSA | Addressing the nurse rostering problem by combining an enhanced HS with a regular hill-climbing algorithm (HC) | CHSA > HS | This hybridization process aided in achieving a balance between exploration and exploitation throughout the search phase. |
| [79], M. Hadwan | HS | Nurse rostering problems | HS | HS obtained reasonable results in addition to other algorithms available in the literature. |
| [80], M. Ayob et al. | Enhanced HS (EHS) | Nurse rostering problems | EHS | EHS showed better results compared to HS. |
| [81], Awadallah MA et al. | HS (improved) | Nurse rostering | HS (improved) | The findings achieved by HS were reasonably equivalent to those acquired by the five INRC2010 winners' approaches. |
| [82], Awadallah MA et al. | Adaptive HS | Nurse scheduling problem | Adaptive HS in comparison with the previously reported results | Adaptive HS was capable of solving nurse scheduling problems. |
| [83], Awadallah MA et al. | Modified HS (MHSA) | Nurse rostering | MHSA | The MHSA built a realistic roster with competitively similar outcomes by using the |

| | | | | International Nurse Rostering Competition 2010 (INRC2010) dataset. |
|---|---|---|---|---|
| [84], Awadallah MA et al. | Hybrid HS (HHSA) | Nurse rostering problem | HHSA | The experimental findings demonstrated that HHSA was capable of producing high-quality solutions within the specified time constraints. |

## 6.9. Miscellaneous

HS has been used in other healthcare systems application areas, such as pressure vessels, diet dwells time optimization, predicting protein structure, medical image enhancement, and finding motifs of biological data. These applications are designed by using either HS or enhanced HS. Table (8) summarizes the main applications of HS and its variants in different areas of healthcare systems.

Table (8): Applications of HS and its variants in different areas of healthcare systems

| Reference, author(s) | Proposed algorithm | Aim | Performance index | Concluding remarks |
|---|---|---|---|---|
| [85], Alomoush AA et al. | HS (improvised) | Designing a simulation software for pressure vessel | HS performed better than other HS hybridized algorithms | The EHS variant provided the best-obtained results both with and without the OBL technique. |
| [86], Firmansyah E et al. | An expert system inspired by HS | Developing an expert method to determine the optimal diet | N/A | The proposed expert system worked well for people with normal body metabolism. |
| [87], Gupta B et al. | Linearly quantile separated histogram equalization-grey relational analysis | Improving the contrast of r mammogram image | The proposed approach works better than state-of-the-art. | The proposed approach showed that basic breast-region segmentation may have provided decent results if the input picture has adequate contrast and can comprehend hidden features. |

| [88], Haridoss R, and Punniyakodi S | Opposition based improved HS (OIHSA) | Reducing the computational time of compressing and enhancing medical images | OIHSA performed well in reducing execution time | The proposed algorithm with Shannon entropy undisputedly performed well |
|---|---|---|---|---|
| [89], Taghipour S et al. | Proposed a methodology | Predicting protein structure | N/A | The proposed methodology established a novel way to predict protein complexes with a high degree of confidence, backed by functional investigations and literature mining. |
| [90], Dongardive J et al. | A novel HS | finding motifs of biological data | N/A | Experimental validation of the suggested approach was performed using sequences of Human Papillomavirus strains acquired from certified and approved sources. |
| [91], Panchal A, and Tom B | HS | Optimizing dwell times for high dose-rate (HDR) prostate brachytherapy | HS > GA | HS was a feasible alternative to currently available algorithms for optimizing HDR prostate brachytherapy. |

## 7. Discussion

As described in the previous Section, HS and its extensions have been used in several healthcare systems applications. Every use of the HS variant has a unique procedure and different performance indicators. Similarly, different HS variants might be deployed for a specific problem, or several issues can be dealt with using the same HS variants. Concerning its performance, HS has been expanded into several versions to solve healthcare problems and has demonstrated excellent performance, with each variant utilizing a unique technique to maintain a high update success rate for HS. However, several forms of HS are thriving in various settings within healthcare systems.

Meanwhile, the HS made an implausible assumption, and its analysis of the update success rate is insufficiently accurate. On this premise, we propose an operational framework of HS variants to summarize their uses and procedures in healthcare problems. The suggested framework is depicted in Figure (7).

The significant expansions of HS in the healthcare systems are the primary HS, HHS, HIS, and multi-objective HS. Each enhancement of HS has been used in several healthcare applications. For example, standard HS and IHS have been used for classification (sound classification and lung nodule classification) and rostering and scheduling problems. In contrast, basic and multi-objective HS have been utilized for smart healthcare unit distributions. Meanwhile, the standard HS has been solely applied to therapeutic medical physics. In addition to that, IHS is the only algorithm deployed for image segmentation and enhancement for COVID-19 detection. On the other hand, heart disease has been predicted using HHS. Other application areas of healthcare systems, such as pressure vessels, diet dwell time optimization, predicting protein structure, medical image enhancement, and finding motifs of biological data, are designed by either HS or improved HS.

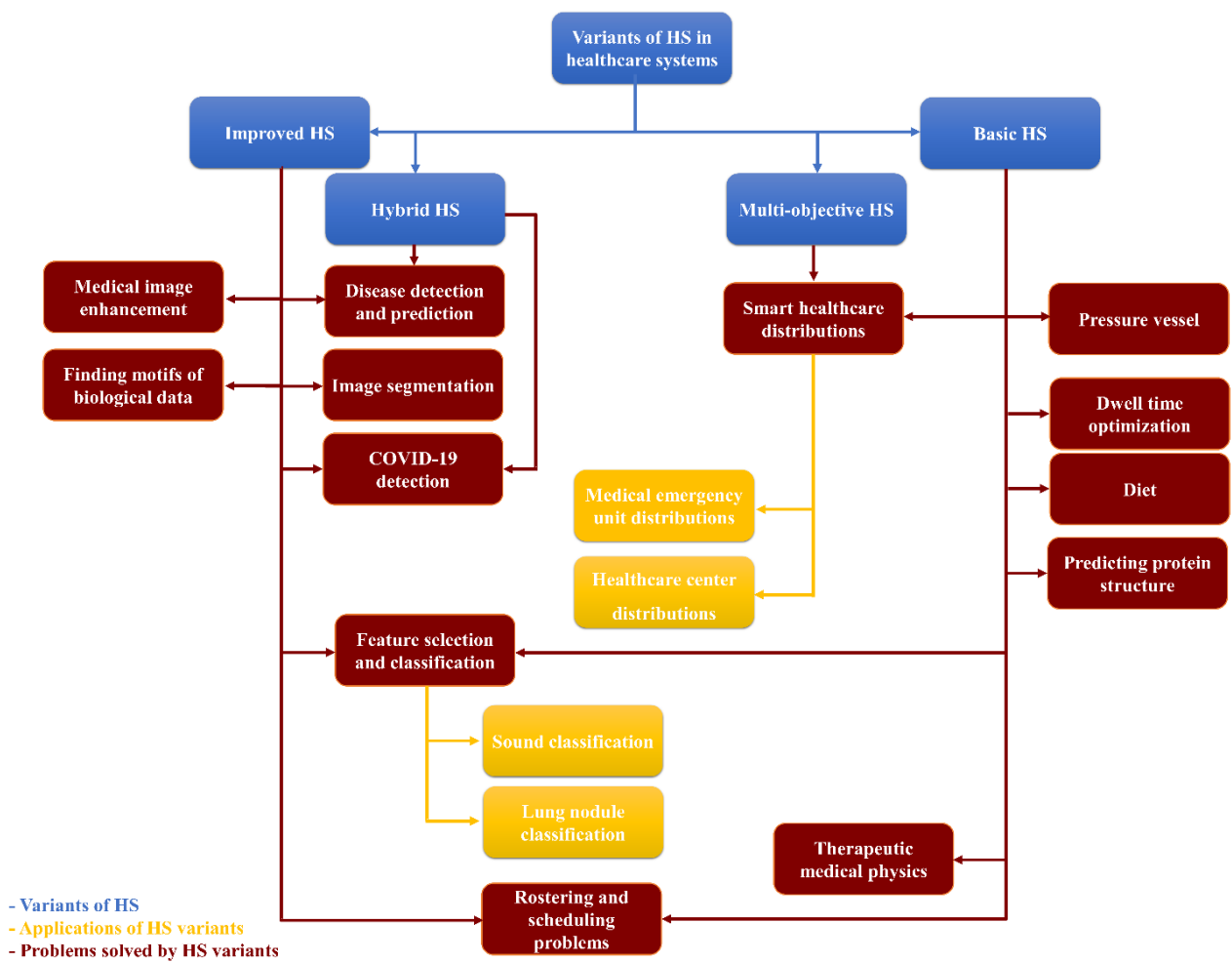

**Figure (7): The proposed operational framework of HS in healthcare systems**

Furthermore, several scholars have made significant efforts to apply HS in healthcare settings, and a system containing the algorithm's operating knowledge is open to other researchers. All these researches proved the popularity of HS, which was gained due to the following advantages:

1) The absence of complex mathematical operations to find the optimum solution;
2) Using a random search, which makes constructive information unnecessary;
3) Simple structure, which enables it to be combined with other meta-heuristic algorithms easily;
4) The speed in finding solutions;
5) Easily applicable in a variety of healthcare applications.

Despite these advantages, it does not prevent the existence of some challenges that mainly revolve around:

1) The static parameters;
2) The freedom to balance local and global search, as it becomes trapped in a local search as extended to numerical applications;
3) Another criticism about the HS is its limited work done mathematically [92,93].

Over the past several decades, scholars have conducted an extensive study on the control parameters of HS. Moreover, several scholars have changed the HS fundamental structure by integrating notions from various metaheuristics. However, there are a few areas in HS that remain unexplored. We believe that the above shortcomings of HS in healthcare applications can be improved in four directions. The most pertinent improvement area of HS is parameter tuning. The most critical phase in every metaheuristic method is parameter tuning. Additional research is necessary to establish the ideal HS parameter setting relevant to healthcare applications. The issue of parameter tuning is seen as an optimization problem. Automatic approaches are needed to determine the optimum parameters for the given health problem. Another exciting enhancement area of HS in health applications is considering the theoretical analysis of HS. The majority of research publications omitted any theoretical consideration of the modification and improvement processes used in HS. There is a need to research the HS and explain why it works better than others. It is necessary to do a theoretical investigation of fitness parameters, structure, and landscapes. The final possible improvement of HS in health systems can be in the scope of multi-objective optimization. Numerous methods for multi-objective harmony search have been developed.

Nonetheless, there is considerable room for expansion in the foreseeable future. Previously used algorithms made use of the idea of the archive. Many other methods and multi-objective operators available in the literature may be used to solve multi-objective issues.

# 8. Conclusion and Future Trends

This review article discussed the HS optimization algorithm, its variations, uses, and implementations in different areas of healthcare systems. As a critical contribution, this study aimed to give a comprehensive assessment of the usage of HS in healthcare systems while also acting as a beneficial resource for potential researchers who want to explore or adopt this approach. Since its inception, the algorithm garnered tremendous attention due to its capability to balance discovery and extraction and its versatility in various research areas due to its simplicity. In recent years, HS had been applied to multiple optimization issues, proving its superiority over other heuristic algorithms and meta-mathematical optimization approaches. The algorithm's continuous evolution and diverse applications to new sorts of healthcare issues demonstrated that HS was an excellent option. As a result, several types of research had been conducted to improve its efficiency. However, other further efforts were possible, including the following:

1) Investigation into how to prevent being trapped in a local solution, since the majority of suggested HS had this issue;
2) Use HS to solve dynamic issues;
3) Create and evaluate methods for an ensemble of HS operators;
4) Propose a novel adaptive approach for parameter updating in the HS;
5) Address more real-world issues, such as the traveling salesman dilemma and time series forecasting;
6) Use HS to tackle difficulties with machine learning;
7) Investigate how well HS performs when confronted with confined issues;
8) Analyze the impact of hybridizing HS with additional EAs.

For additional research in the future, HS can be hybridized with several algorithms for healthcare problems to further validate its efficiency, such as the backtracking search optimization algorithm [94–96], the variants of evolutionary clustering algorithm star [97–100], chaotic sine cosine firefly algorithm [101], shuffled frog leaping algorithm [102] and hybrid artificial intelligence algorithms [103]. Furthermore, HS can be applied to more complex and real-world applications to explore more deeply the advantages and drawbacks of the algorithm or improve its efficiencies, such as engineering application problems [101], wind speed prediction [104–110], traffic flow prediction [111], laboratory management [112], e-organization and e-government services [113], online analytical processing [114], web science [115], and the Semantic Web ontology learning [116].

**Acknowledgments**

Some special thanks go to the University of Kurdistan Hewler and Kurdistan Institution for Strategic Studies and Scientific Research for their help and willingness to conduct this review.

**Compliance with Ethical Standards**

**Conflict of interest:** None.

**Funding:** No funding is applicable.